\documentclass[conference]{IEEEtran}
\IEEEoverridecommandlockouts
\usepackage{cite}
\usepackage{amsmath,amssymb,amsfonts}
\usepackage{algorithmic}
\usepackage{graphicx}
\usepackage{textcomp}
\usepackage{xcolor}
\begin{document}

\title{Unified State Representation Learning under Data Augmentation\\
}

\author{\IEEEauthorblockN{1\textsuperscript{st} Taylor Hearn}
\IEEEauthorblockA{\textit{College of Computing} \\
\textit{Georgia Institute of Technology}\\
Atlanta, Georgia \\
thearn6@gatech.edu}
\and
\IEEEauthorblockN{1\textsuperscript{st} Sravan Jayanthi}
\IEEEauthorblockA{\textit{College of Computing} \\
\textit{Georgia Institute of Technology}\\
Atlanta, Georgia \\
sjayanthi@gatech.edu}
\and
\IEEEauthorblockN{2\textsuperscript{nd} Sehoon Ha}
\IEEEauthorblockA{\textit{College of Computing} \\
\textit{Georgia Institute of Technology}\\
Atlanta, Georgia \\
sehoonha@gatech.edu}
}

\maketitle

\begin{abstract}
The capacity for rapid domain adaptation is important to increasing the applicability of reinforcement learning (RL) to real world problems. Generalization of RL agents is critical to success in the real world, yet zero-shot policy transfer is a challenging problem since even minor visual changes could make the trained agent completely fail in the new task. We propose USRA: \underline{U}nified \underline{S}tate \underline{R}epresentation Learning under Data \underline{A}ugmentation, a representation learning framework that learns a latent unified state representation by performing data augmentations on its observations to improve its ability to generalize to unseen target domains. We showcase the success of our approach on the DeepMind Control Generalization Benchmark for the Walker environment and find that USRA achieves higher sample efficiency and 14.3\% better domain adaptation performance compared to the best baseline results.

\end{abstract}

\section{Introduction}
% Can cut from the introduction
Latent Unified State Representation \cite{xing2021domain} (LUSR) is a representation learning technique used for zero-shot domain adaptation from a source task to related target tasks (i.e. same action space and similar transitions/rewards). In zero-shot learning, the agent is only trained on the source domain; while restrictive, this approach is applicable to real life problems such as autonomous driving under different weather conditions. 
% \footnote{
% We note that this problem formulation differs from that of domain randomization since we only have access to a single domain and we want to generalize to multiple unseen domains. Domain randomization violates the zero-shot constraint and its complexity scales significantly with the number of domain variations.}
% Note that domain randomization uses a different problem statement, since it allows RL training on multiple domains (violates the zero-shot constraint), so it cannot be fairly compared to LUSR. 

LUSR trains a cycle-consistent Variational Autoencoder (VAE) \cite{DBLP:journals/corr/abs-1804-10469} that outputs a domain general representation, with features like the shape of the road, and a domain specific representation, with features like the background color, by learning from unlabeled images taken from different tasks. The encoder is then frozen and the RL agent is trained using the learned domain general representations.

We hypothesize that we can improve LUSR by fine-tuning the encoder during the training of the RL agent rather than freezing it. We propose 
\textbf{USRA}, \underline{U}nified \underline{S}tate \underline{R}epresentation Learning under Data \underline{A}ugmentation, which is a technique that trains an encoder to learn a generalizable state representation using a pretraining phase followed by finetuning during policy learning, as shown in Figure 1. We challenge the encoder to learn online from observations collected by the policy to encourage a more useful image embedding than one only trained on random observations.

\begin{figure}[th]
\begin{center}
\includegraphics[width=9cm]{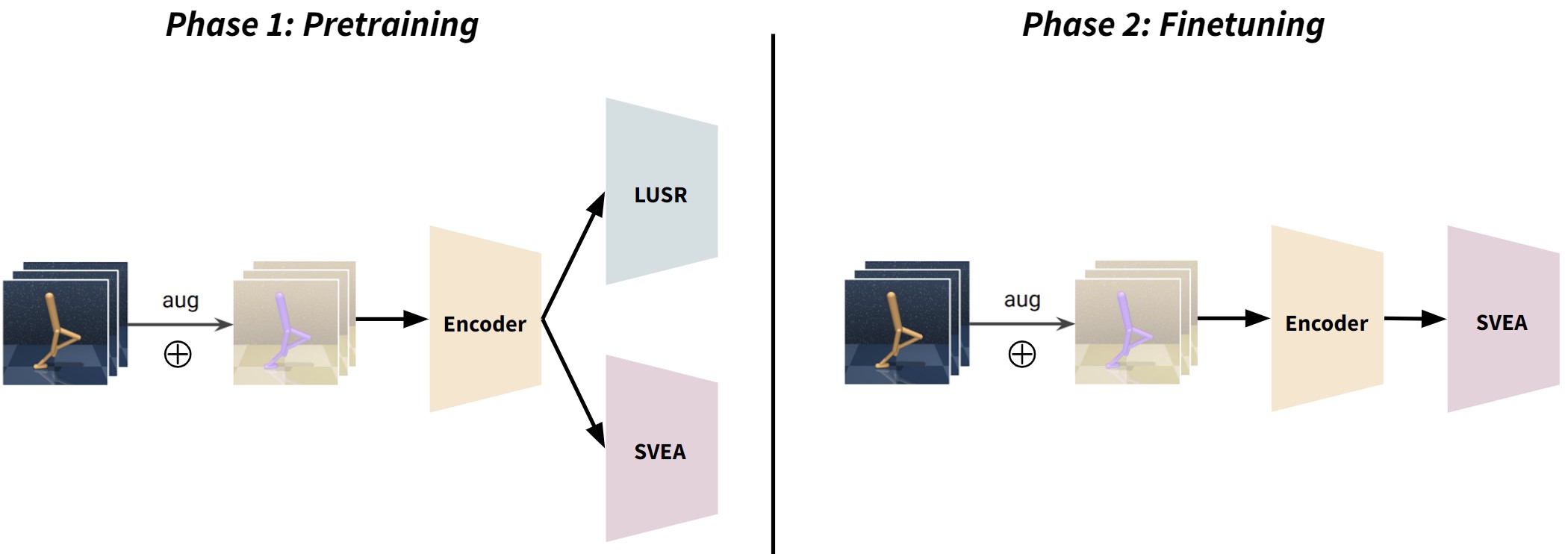}
\caption{Overview of the architecture of USRA during the pretraining and finetuning phase. USRA leverages image augmentations to train the encoder.  $\oplus$ is a concatenation operation for input frames.}
\end{center}
\label{figusra}
\end{figure}

Our intuition is that seeing more varied (rather than random) observations from different areas of the state space that are explored during policy training can improve the encoder's learned representations. This should allow the representation to incorporate information that is relevant to achieving a higher return rather than just for image reconstruction. 
To enable fine-tuning without loss of generality, we add an auxiliary objective during agent training called SVEA (Stabilized Q-Value Estimation under Augmentation) \cite{NEURIPS2021_1e0f65eb}. This method is a domain generalization technique that adds a consistency loss term between the estimated Q-value of the latent representation of an augmented version of the frame and the target Q-value of the original frame. The augmented frames are used only to compute the SVEA loss while policy training occurs with the original non-augmented frames.

We evaluate USRA on a domain generalization benchmark called the DeepMind Control Generalization Benchmark (DMControl-GB) \cite{hansen2021generalization}. This benchmark is based on continuous control tasks from the MuJoCo physics engine \cite{todorov2012mujoco} and are perturbed with random colors or video backgrounds during evaluation time. We compare our method on the Walker task to the baselines of LUSR and SVEA, and find USRA outperforms all methods with better sample efficiency, asymptotic performance, and generalization success.

Our contribution in this paper is three-fold:

\begin{enumerate}
\item We propose USRA, a unified state representation learning technique that decomposes an image into a domain specific and domain general embedding and improves the encoder through an auxiliary Q-value estimation objective on augmented images.
\item We find that USRA has better sample efficiency than either of the baselines LUSR or SVEA.
\item USRA has a better generalization capacity than best baselines as it achieves 22.6\% higher returns on an unseen task that randomly varies the background video of Walker and 21.2\% on a harder unseen task that overlays a video on the entire environment. 

\end{enumerate}

\section{Related Work}
Self-Supervised reinforcement learning has been shown to improve the data efficiency of reinforcement learning, particularly in visual domains \cite{DBLP:journals/corr/abs-2004-04136, DBLP:journals/corr/abs-2004-13649, DBLP:journals/corr/abs-2007-05929, DBLP:journals/corr/abs-2106-04152}. Some techniques that involve training encoders include applying auxiliary losses on data augmented to input frames to realize a consistent latent representation \cite{DBLP:journals/corr/abs-2004-04136, DBLP:journals/corr/abs-2004-13649}. Other techniques involve modeling a forward and reverse dynamics model to ensure the latent representation encodes temporal information \cite{DBLP:journals/corr/abs-2007-05929, DBLP:journals/corr/abs-2106-04152}.  However, these approaches rely on the availability of multiple source domains for training and the complexity of this approach scales with the number of variations. Instead of learning a policy with generalization capability directly, our work (USRA) focuses on the generalization of state representations.

\section{Method}
\subsection{Markov Decision Process (MDP)}
A Markov Decision Process $M$ is described as a 6-tuple $\langle \mathbb{S},\mathbb{A},R,T,\gamma,\rho_0\rangle$. $\mathbb{S}$ and $\mathbb{A}$ represent the state and the action space, respectively. $R$ denotes the reward function, meaning the environment provides $R(s)$ reward at state $s$. $T(s^\prime\vert s,a)$ is the probability of transitioning into state $s^\prime$ after taking action $a$ in state $s$. $\gamma\in(0,1)$ is the temporal discount factor, controlling the trade-off between instantaneous reward and future rewards. $\rho_0$
% : \mathbb{S}\rightarrow \mathbb{R}$ 
denotes the initial state probability and policy $\pi$
% : \mathbb{S}\times\mathbb{A}\rightarrow \mathbb{R}$
is the probability of choosing an action given a state. 
% An infinite-horizon MDP alternates between policy executions and environment transitions: $s_1\sim \rho_0(\cdot),\ a_t\sim\pi(\cdot\vert s_t),\ s_{t+1}\sim T(\cdot\vert s_t,a_t),\ t=1,2,\cdots$. 
% The standard objective for RL is to find the optimal policy that maximizes the discounted return, $\pi^*=\arg\max_\pi\mathbb{E}_{\tau\sim\pi}\left[\sum_{t=1}^\infty{\gamma^{t-1}R(s_t)}\right]$, and $\tau\sim\pi$ means the trajectory $\tau$ is induced by the policy $\pi$ and implicitly contains the state-action sequence: $\{s_1,a_1,s_2,a_2,\cdots\}$. We instead consider the maximum entropy objective and leverage the Soft Actor Critic \cite{DBLP:journals/corr/abs-1801-01290} method as our underlying policy learning method in USRA.

\subsection{Latent Unified State Representation (LUSR)}
The LUSR method transforms the state space $\mathbb{S}$ of an MDP from an agent's raw observation state space $\mathbb{S}^o$ to a latent state space $\mathbb{S}^z$ through the function $f: \mathbb{S}^o \rightarrow \mathbb{S}^z$. LUSR decomposes the latent state space into disjoint domain-specific and domain-general features $\mathbb{S}^z = (\widehat{\mathbb{S}^z}, \overline{\mathbb{S}^z})$ ($\widehat{\mathbb{S}^z} = $ domain specific, $\overline{\mathbb{S}^z}$ = domain general). Intuitively, a domain-general feature is one that is useful across similar domains (like the agent's position on the screen), while a domain-specific feature is particular to one domain (like the background color).

LUSR employs a Cycle-Consistent VAE \cite{DBLP:journals/corr/abs-1804-10469} to disentangle the domain-general and domain-specific features. In the forward cycle of the Cycle-Consistent VAE, for two observation states $s_1^o, s_2^o \in \mathbb{S}^o$ from the same domain, $Enc(s_1^o)=\widehat{s_1^z},\overline{s_1^z}$ and $Enc(s_2^o)=\widehat{s_2^z},\overline{s_2^z}$, LUSR swaps the domain-specific embeddings and reconstructs the image using the decoder such that $Dec(\widehat{s_2^z},\overline{s_1^z})\approx s_1^o$ and $Dec(\widehat{s_1^z},\overline{s_2^z})\approx s_2^o$. The loss, with encoder $f_{\theta}$ and decoder $d_{\theta}$ described in Equation \ref{equation1}, ensures that the domain-general encoding contains sufficient information to reconstruct the original input observation.

\begin{equation}
  \mathcal{L}_{forward} = - \mathbb{E}_{d_{\theta}(\widehat{s^z},\overline{s^z}|s^o)} \log f_{\theta}(s^o|\widehat{s^z},\overline{s^z})
  \label{equation1}
\end{equation}

In the reverse cycle of the VAE, a randomly sampled $\overline{s^z}$ is transformed with two domain-specific embeddings $\widehat{s_1^z},\widehat{s_2^z}$ to encourage the encoder to recover the latent domain-general embedding
$Enc(Dec(\widehat{s_1^z},\overline{s^z})) = \widehat{s_x^z},\overline{s_x^z}$ and $Enc(Dec(\widehat{s_2^z},\overline{s^z})) = \widehat{s_y^z},\overline{s_y^z}$. The loss, described in Equation \ref{equation2} seeks to enforce $\overline{s_x^z} \approx \overline{s_y^z}$ to encourage a domain-general embedding that is invariant across domains.

\begin{equation}
  \mathcal{L}_{reverse} =  \mathbb{E}_{\overline{s^z} \sim f(\overline{s^z})} || d_{\theta}(f_{\theta}(\widehat{s_1^z},\overline{s^z})) - d_{\theta}(f_{\theta}(\widehat{s_2^z},\overline{s^z})) ||_1
  \label{equation2}
\end{equation}

\subsection{Stabilized Q-Value Estimation under Augmentation (SVEA)}
SVEA estimates the $Q_{\theta}$ function of the MDP using an encoder, $f_{\theta}$, where the  predicted Q-value is defined as $q_t \triangleq Q_{\theta} (p_{\theta}(s_t, a_t))$. The target state-action value function is $Q_{\psi}$ where $\psi$ is the exponential moving average of $\theta$ defined in Equation \ref{equation3}:

\begin{equation}
  \psi_{n+1} \leftarrow (1-\zeta) \psi_n + \zeta \theta_n
  \label{equation3}
\end{equation}

for iteration step $n$ and momentum coefficient $\zeta \in [0,1]$. SVEA performs stable Q-value estimation by updating the target state-action value function according to a temporal difference objective defined in Equation \ref{equation4}.

\begin{equation}
  q_t^{tgt} \triangleq r(s_t, a_t) + \gamma \max_{a_t'} Q_{\psi}^{tgt} (f_{\psi}^{tgt}(s_{t+1}, a_t'))
  \label{equation4}
\end{equation}

SVEA leverages a collection of random image augmentations to transform an observation $s_t^{aug} = v(s_t), v \sim \mathcal{V}$. The original images and the augmented images are used in the $Q$-value estimation loss to encourage the estimated $Q$-value of both types of images to align with the target $Q$-value as described in Equation \ref{equation5}.

\begin{multline}
  \mathcal{L}_{SVEA} =  \mathbb{E}_{s_t, a_t, s_{t+1} \sim M} [ ||Q_{\theta}(f_{\theta}(s_t), a_t) - q_t^{tgt} ||_2^2 + \\ ||Q_{\theta}(f_{\theta}(s_t^{aug}), a_t) - q_t^{tgt} ||_2^2 ]
  \label{equation5}
\end{multline}

\subsection{Unified State Representation Learning under Augmentation (USRA)}

USRA learns a unified state representation $f: \mathbb{S}^o \rightarrow \mathbb{S}^z$ from the raw observation space $\mathbb{S}^o$ to a latent embedding space $\mathbb{S}^z$. 
The learned encoder identifies a unified state representation through a projection $g_\theta: S^z \rightarrow \widehat{S^z}, \overline{S^z}$ which composes to identify a domain specific and domain general encoding $g_\theta(f_\theta(s)) = \widehat{s^z},\overline{s^z}$. The auxiliary loss in $Q$-value estimation leverages a different projection head $Q_{\theta}$ to obtain the $Q$-value estimate $Q_\theta(f_\theta(s))$. These two networks are trained simultaneously during the first phase $P^{(1)}$ and then $Q_{\theta}$ is fine-tuned during the second phase $P^{(2)}$.

During a initial pre-training phase $P^{(1)}$, state-action pairs collected using a random policy, $((s_0, a_0), ..., (s_n, a_n)) \sim \dot{\pi}$, USRA leverages a non-negative linear combination of the Cycle-Consistency loss as well as $Q$-value estimation objective as shown in Equation \ref{equation6},

\begin{equation}
  \mathcal{L}_{USRA} = \beta_1 (\mathcal{L}_{forward} + \mathcal{L}_{reverse}) + \beta_2 \mathcal{L}_{SVEA}
  \label{equation6}
\end{equation} 

with weights $\beta_1$ and $\beta_2$ on each loss. Then during the agent learning phase $P^{(2)}$, USRA will enforce the temporal difference loss $ \mathcal{L}_{SVEA}$ on augmented and non-augmented states $s_t, s_t^{aug} \sim v(s_t)$ collected from the policy during training $s_t \sim \pi_{\theta}$. 
% We propose two distinct variations of USRA during the second learning phase: 

% \begin{enumerate}
% \item USRA leverages the SVEA auxiliary loss to fine-tune the encoder,
% \item Both the SVEA auxiliary and LUSR objective are used to fine-tune the encoder during $P^{(2)}$.
% \end{enumerate}
% We call the first variation Alpha 1 and the second Alpha 2.

\begin{table}[t]
\caption{Comparison of adaptation performance on Walker domains.}
\begin{center}
\begin{tabular}{|c|c|c|c|c|c|}
\hline
\cline{2-4} 
\textbf{} & \textbf{\textit{Train}} & \textbf{\textit{Color}} & \textbf{\textit{Color}} & \textbf{\textit{Video}} & \textbf{\textit{Video}} \\
\textbf{} & \textbf{} & \textbf{\textit{(Easy)}} & \textbf{\textit{(Hard)}} & \textbf{\textit{(Easy)}} & \textbf{\textit{(Hard)}} \\
\hline
USRA & \textbf{949} & \textbf{949} & \textbf{948} & \textbf{862} & \textbf{245}\\
SVEA & 892 & 888 & 871 & 703 & 202 \\
LUSR & 374 & 273 & 150 & 165 & 43 \\
\hline
% \multicolumn{4}{l}{$^{\mathrm{a}}$Sample of a Table footnote.}
\end{tabular}
\label{tab1}
\end{center}
\end{table}

Data collection is an important concern for the training of the Cycle-Consistent VAE, since it requires gathering informative observations from multiple domains without a trained policy. LUSR required a large dataset of observations collected offline, randomly sampled from the source domain and a subset of the target domains (called the seen target domains). USRA overcomes this limitation by leveraging just observations from the source domain and applies augmentations on those observations to imitate the concept of target domains without needing additional samples from such domains.

\section{Results}
We design experiments to validate that the proposed
framework, USRA, can more efficiently learn a well performing policy generalize successfully to unseen target domains. Particularly, our goal is to investigate the following questions:
\begin{enumerate}
\item Can USRA learn a better performing policy with fewer samples compared to baseline approaches?
\item Does USRA generalize better under challenging distribution shifts than baseline approaches?
\end{enumerate}

We will be evaluating USRA with these questions in mind on the Walker environment from the DMControl Generalization Benchmark. The goal for the Walker task is for a 2D Bipedal Humanoid to move to forward as quickly as possible by learning a stable and effective gait. We trivially set the value $\beta_1=\beta_2$ for the USRA loss. The augmentation that USRA uses in order to modify observations as part of the $Q$-value auxiliary loss is a random convolution $c \sim \mathcal{C}$ on each observation such that $s_t^{aug} = c(s_t)$. The number of frames used for $P^{(1)}$ pre-training of the encoder is 1000 frames collected from a random policy with random initialization. The learning rate for the encoder, policy, and all projection networks is 0.001. The batch size for the LUSR loss is 16 and for SVEA fine-tuning is 128.

\subsection{Comparison of Adaptation Algorithms}
We compare USRA with the baseline methods LUSR and SVEA. We train each method for 2,000 episodes with an episode length of 1,000. We take the best performing policy learned from each method and find in Table \ref{tab1} that USRA outperforms both baselines.

We find that after training, USRA learns a policy that has 6.3\% higher returns in the source domain compared to the next best method, SVEA. More importantly, USRA outperforms the other baselines in the domain adaptation task on the unseen target domains. In the Color easy domain, where the background and platform color are changed, USRA performs 6.9\% better than SVEA. In the Color hard domain, where additionally the Walker agent's color is modified, USRA achieves 8.8\% better generalization performance. With in the Video easy domain, where a random video is played in place of a static background, USRA learns an agent that 22.6\% better returns. For the Video hard domain, where the platform is also removed and a video is overlayed on the entire backdrop, USRA outperforms SVEA by 21.2\%. 

The training and evaluation curves of USRA and the baselines are shown in Figure 2.

\begin{figure}[t]
\begin{center}
\includegraphics[width=9cm]{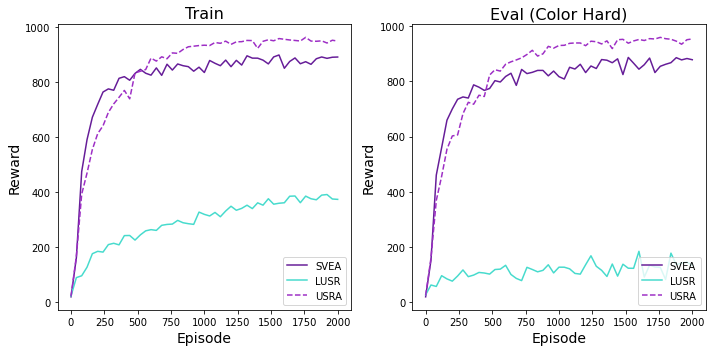}
\caption{Training curve of average train and color (hard) eval reward for LUSR, SVEA, and USRA.}
\end{center}
\label{fig1}
\end{figure}

For 500 episodes, USRA has similar sample efficiency as the best baselines, SVEA, and for 1000 episodes, USRA has better sample efficiency. The asymptotic performance of USRA in both the training and evaluation domains exceeds that of the baselines suggesting that USRA learns an encoder that facilitates the rapid training of a high-performing policy. 

\subsection{USRA Ablations}
\begin{figure}[t]
\begin{center}
\includegraphics[width=9cm]{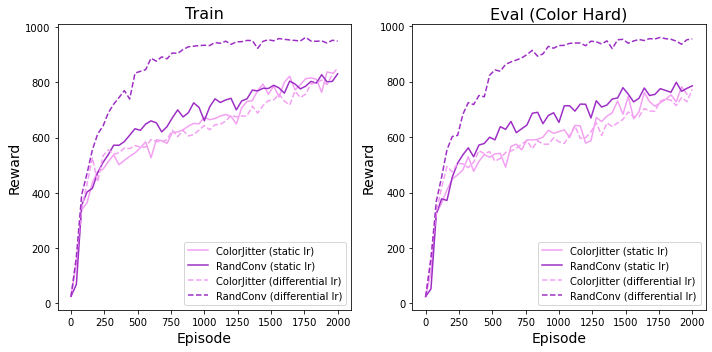}
\caption{Training curve of average train and color (hard) eval reward for different variations of USRA.}
\end{center}
\label{fig2}
\end{figure}

We have ablated two different design choices for USRA. The first ablation is of the type of augmentation that USRA uses in the encoder pretraining phase to create different domains. We tried two different types of augmentations: random convolution and color jitter. Our random convolution augmentations create a 3x3 filter with random weights and apply it to an image, while our color jitter augmentations significantly vary the hue and slightly vary the brightness, contrast, and saturation of an image. As can be seen in \ref{fig2}, random convolutions (\textit{RandConv}) lead to significantly higher sample efficiency and asymptotic performance. We hypothesize that this could be due to random convolution being a much stronger augmentation than color jittering. Having an encoder that has learned to extract domain-general features from more strongly augmented frames could allow it to adapt to a wider range of domains during the fine-tuning phase.

The second ablation is whether to lower the learning rate of the encoder by 10x after the pretraining step. It is common practice to lower the learning rate of the earlier layers when finetuning a pretrained model to prevent catastrophic forgetting. In Figure 3, \textit{static lr} refers to the uniform learning rate case while \textit{differential lr} refers to the case where the encoder has a lower learning rate after pretraining. We found that lowering the learning rate of the encoder significantly improved sample efficiency and the final reward values for the random convolution variant, but had no measurable effect on the color jitter variant. This could be because the pretrained color jitter encoder is not significantly more useful than a random initialization, so catastrophic forgetting is not an issue for it. On the other hand, the pretrained random convolution encoder seems to have learned something more useful using the same reasoning.

We picked the most successful configuration, USRA with random convolutions and differential learning rate, to compare to the baselines. This is the version of USRA that we are referring to whenever we mention USRA without qualification. 

\subsection{Analysis}
Our proposed method, USRA, accomplishes both goals we set out to investigate in our DMControl Generalization Walker experiments. USRA is able to demonstrate better sample efficiency after 500 episodes compared to SVEA. This is because USRA is able to leverage pretraining for the encoder, and thus can learn to identify domain general and domain specific features of observations. Then, during the finetuning stage, it can incorporate new information from policy rollouts to improve the learned representations under the shift of state marginal distribution when the policy is exploring. 

Not only does it show better sample efficiency, but USRA further exceeds expectations by demonstrating better asymptotic performance on the training and evaluation domains compared to the baseline approaches. LUSR shows poor asymptotic performance since its training dataset is limited to observations that are seen by a random policy. Therefore, much of the state space is unseen after LUSR freezes the encoder and performs policy training. On the other hand, SVEA lacks a cycle-consistency loss, so it does not regularize the mapping from the observation to the learned embeddings such that there exists a reverse transformation that can reconstruct the original image. USRA resolves both of these issues by learning a representation that is transitive while also finetuning that representation along unseen ranges of state space to maximize performance.

USRA also accomplishes the second challenge of domain adaptation by successfully generalizing under challenging distribution shifts. This suggests that the LUSR objective (in USRA's pretraining) of identifying domain-general and domain-specific embeddings improves the adaptation capacity, since the encoder is encouraged to identify meaningful features and disregard domain specific features such as the platform coloring or the background.

\section{Conclusion and Future Work}
We present USRA, Unified State Represention under Augmentation, which successfully learns generalizable state representations through encoder pretraining and finetuning using data augmentations. USRA demonstrates higher sample efficiency than the baseline methods and succeeds at the problem of zero-shot domain adaptation on unseen domains. USRA builds upon the previous technique, LUSR, which enables zero-shot adaptation by training a cycle-consistent VAE on random observations from the source and seen target domains, using the learned encoder as a latent representation for RL training on the source domain. USRA leverages the stable $Q$-value estimation technique presented by SVEA to fine-tune the learned representations while exploring new parts of the state space during agent training.

Since our method can perform zero shot transfer, for future work we want to study how well USRA performs in out-of-distribution domains, where the action space $\mathbb{A}_d$ may be different. We can use our learned encoder $p_{\theta}$ for few shot training for a new policy $\pi_d: \mathbb{S} \rightarrow \mathbb{A}_d$. Additionally, this method could potentially be effective in few shot learning for sim-to-real transfer. The recovered encoder can identify robust and generalizable representations that can adapt to the domain shift seen in real world robot scenarios. The advantage of this approach is that it will only require a few real-world images to be able to be able to deploy on a robot.

We also want to study how stronger augmentations could improve the learned representation. Currently, USRA only uses either color jitter or random convolutions. Vision Transformers have achieved impressive results in downstream tasks in computer vision, and are commonly used with very strong augmentations like CutMix and MixUp, could allow us to use even stronger augmentations than random convolution and and thus encourage better generalization in the learned representations.

Domain generalization is another way of training an agent to be robust to minor domain variations. It would be useful to compare these results to a baseline of domain generalization, as well as investigate whether USRA could be combined with domain randomization in some form. Perhaps USRA could rapidly train a capable agent, then domain randomization could gradually be introduced in a curriculum learning fashion to increase the asymptotic performance on large domain shifts (like the video background domains).

\bibliography{report.bib}{}

\begin{thebibliography}{1}
\providecommand{\url}[1]{#1}
\csname url@samestyle\endcsname
\providecommand{\newblock}{\relax}
\providecommand{\bibinfo}[2]{#2}
\providecommand{\BIBentrySTDinterwordspacing}{\spaceskip=0pt\relax}
\providecommand{\BIBentryALTinterwordstretchfactor}{4}
\providecommand{\BIBentryALTinterwordspacing}{\spaceskip=\fontdimen2\font plus
\BIBentryALTinterwordstretchfactor\fontdimen3\font minus
  \fontdimen4\font\relax}
\providecommand{\BIBforeignlanguage}[2]{{%
\expandafter\ifx\csname l@#1\endcsname\relax
\typeout{** WARNING: IEEEtran.bst: No hyphenation pattern has been}%
\typeout{** loaded for the language `#1'. Using the pattern for}%
\typeout{** the default language instead.}%
\else
\language=\csname l@#1\endcsname
\fi
#2}}
\providecommand{\BIBdecl}{\relax}
\BIBdecl

\bibitem{xing2021domain}
J.~Xing, T.~Nagata, K.~Chen, X.~Zou, E.~Neftci, and J.~L. Krichmar, ``Domain
  adaptation in reinforcement learning via latent unified state
  representation,'' \emph{arXiv preprint arXiv:2102.05714}, 2021.

\bibitem{DBLP:journals/corr/abs-1804-10469}
\BIBentryALTinterwordspacing
A.~H. Jha, S.~Anand, M.~Singh, and V.~S.~R. Veeravasarapu, ``Disentangling
  factors of variation with cycle-consistent variational auto-encoders,''
  \emph{CoRR}, vol. abs/1804.10469, 2018. [Online]. Available:
  \url{http://arxiv.org/abs/1804.10469}
\BIBentrySTDinterwordspacing

\bibitem{NEURIPS2021_1e0f65eb}
N.~Hansen, H.~Su, and X.~Wang, ``Stabilizing deep q-learning with convnets and
  vision transformers under data augmentation,'' in \emph{Advances in Neural
  Information Processing Systems}, vol.~34, 2021, pp. 3680--3693.

\bibitem{hansen2021generalization}
N.~Hansen and X.~Wang, ``Generalization in reinforcement learning by soft data
  augmentation,'' in \emph{2021 IEEE International Conference on Robotics and
  Automation (ICRA)}.\hskip 1em plus 0.5em minus 0.4em\relax IEEE, 2021, pp.
  13\,611--13\,617.

\bibitem{todorov2012mujoco}
E.~Todorov, T.~Erez, and Y.~Tassa, ``Mujoco: A physics engine for model-based
  control,'' in \emph{2012 IEEE/RSJ international conference on intelligent
  robots and systems}.\hskip 1em plus 0.5em minus 0.4em\relax IEEE, 2012, pp.
  5026--5033.

\bibitem{DBLP:journals/corr/abs-2004-04136}
\BIBentryALTinterwordspacing
A.~Srinivas, M.~Laskin, and P.~Abbeel, ``{CURL:} contrastive unsupervised
  representations for reinforcement learning,'' \emph{CoRR}, vol.
  abs/2004.04136, 2020. [Online]. Available:
  \url{https://arxiv.org/abs/2004.04136}
\BIBentrySTDinterwordspacing

\bibitem{DBLP:journals/corr/abs-2004-13649}
\BIBentryALTinterwordspacing
I.~Kostrikov, D.~Yarats, and R.~Fergus, ``Image augmentation is all you need:
  Regularizing deep reinforcement learning from pixels,'' \emph{CoRR}, vol.
  abs/2004.13649, 2020. [Online]. Available:
  \url{https://arxiv.org/abs/2004.13649}
\BIBentrySTDinterwordspacing

\bibitem{DBLP:journals/corr/abs-2007-05929}
\BIBentryALTinterwordspacing
M.~Schwarzer, A.~Anand, R.~Goel, R.~D. Hjelm, A.~C. Courville, and P.~Bachman,
  ``Data-efficient reinforcement learning with momentum predictive
  representations,'' \emph{CoRR}, vol. abs/2007.05929, 2020. [Online].
  Available: \url{https://arxiv.org/abs/2007.05929}
\BIBentrySTDinterwordspacing

\bibitem{DBLP:journals/corr/abs-2106-04152}
\BIBentryALTinterwordspacing
T.~Yu, C.~Lan, W.~Zeng, M.~Feng, and Z.~Chen, ``Playvirtual: Augmenting
  cycle-consistent virtual trajectories for reinforcement learning,''
  \emph{CoRR}, vol. abs/2106.04152, 2021. [Online]. Available:
  \url{https://arxiv.org/abs/2106.04152}
\BIBentrySTDinterwordspacing

\end{thebibliography}
\bibliographystyle{IEEEtran}

\end{document}